\documentclass{bmvc2k}

\title{Spatio-Temporal Learnable Proposals for End-to-End Video Object Detection}

\addauthor{Khurram Azeem Hashmi}{khurram_azeem.hashmi@dfki.de}{1,2}
\addauthor{Didier Stricker}{didier.stricker@dfki.de}{1,2}
\addauthor{Muhammamd Zeshan Afzal}{muhammamd_zeshan.afzal@dfki.de}{1,2}

\addinstitution{
German Research Centre for Artificial \\
Intelligence (DFKI) \\
Kaiserslautern, Germany
}
\addinstitution{
Technical University of Kaiserslautern\\
Germany
}
\runninghead{Hashmi et al}{Spatio-Temporal Learnable Proposals for End-to-End VOD}

\def\etal{\emph{et al}\bmvaOneDot}

\begin{document}

\maketitle

\begin{abstract}
This paper presents the novel idea of generating object proposals by leveraging temporal information for video object detection.~The feature aggregation in modern region-based video object detectors heavily relies on learned proposals generated from a single-frame RPN. This imminently introduces additional components like NMS and produces unreliable proposals on low-quality frames.~To tackle these restrictions, we present SparseVOD, a novel video object detection pipeline that employs Sparse R-CNN to exploit temporal information.~In particular, we introduce two modules in the dynamic head of Sparse R-CNN. First, the Temporal Feature Extraction module based on the Temporal RoI Align operation is added to extract the RoI proposal features.~Second, motivated by sequence-level semantic aggregation, we incorporate the attention-guided Semantic Proposal Feature Aggregation module to enhance object feature representation before detection.~The proposed SparseVOD effectively alleviates the overhead of complicated post-processing methods and makes the overall pipeline end-to-end trainable.~Extensive experiments show that our method significantly improves the single-frame Sparse R-CNN by 8\%-9\% in mAP.~Furthermore, besides achieving state-of-the-art 80.3\% mAP on the ImageNet VID dataset with ResNet-50 backbone, our SparseVOD outperforms existing proposal-based methods by a significant margin on increasing IoU thresholds (IoU > 0.5).
\end{abstract}

\section{Introduction}
\label{sec:intro}

Video Object Detection~(VOD) aims to localize and classify objects in a series of subsequent video frames. Recent efforts in video object detection demonstrate that exploiting feature aggregation of temporal information~\cite{zhu2017deep, zhu2017flow, feichtenhofer2017detect, wang2018fully, wu2019sequence, han2020mining, gong2021temporal, han2021class, cui2021tf, hua2021temporal, han2020exploiting} produce superior performance than leveraging temporal information at the post-processing stage~\cite{han2016seq, kang2016object, kang2017object, sabater2020robust}. The former approaches mainly enhance the target frame feature representation through aggregating features from neighbouring frames or an entire video clip by designing a specific module, thereby boosting detection results.~The majority of these works~\cite{wu2019sequence, hua2021temporal, jiang2020learning, han2020mining, shvets2019leveraging} employ two-stage detectors such as Faster R-CNN~\cite{ren2015faster} or R-FCN~\cite{dai2016r} to design their VOD pipelines. 

\begin{figure}
\includegraphics[width=12.5cm]{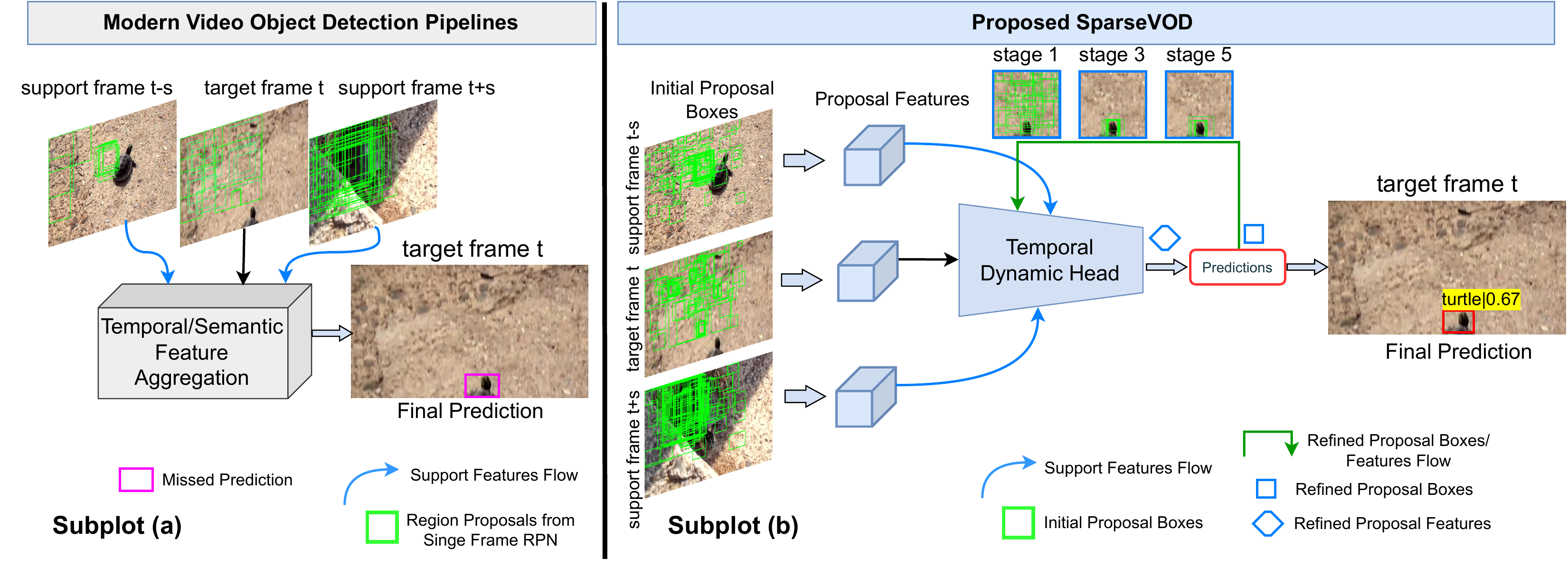}
\caption{Comparison between previous region-based and the proposed video object detection methods. Subplot (a): Despite the temporal feature aggregation from support frames \textit{t-s} and \textit{t+s} in existing proposal-based approaches, the detector overlooks \textit{Turtle} in \textcolor{violet}{pink} on a low-quality target frame \textit{t} due to unreliable region proposals. Subplot (b): To tackle such limitations, we propose a novel SparseVOD framework that iteratively refines region proposals through leveraging spatio-temporal feature aggregation prior to final detection. The Temporal Dynamic Head is explained in Fig.~\ref{fig:pipeline}.}
\label{fig:problem_definition}
\vspace{-15pt}
\end{figure}

Albeit the enormous success, it is important to highlight that the temporal feature aggregation scheme of all prior region-based VOD methods~\cite{wu2019sequence, han2021class, han2020mining, hua2021temporal, jiang2020learning, cui2021tf} heavily relies on object proposals from RPN~\cite{ren2015faster} trained without any temporal information. Consequently, these methods suffer from several underlying restrictions. First, they require an additional step of NMS~\cite{neubeck2006efficient} at the beginning to perform dense-to-sparse matching of hand-crafted anchors, making the overall VOD pipeline not end-to-end optimizable. Second, on low-quality frames (appearance deterioration), the generated object proposals are unreliable, leading to ineffective temporal feature aggregation. 
Fig.~\ref{fig:problem_definition}\textcolor{red}{(a)} illustrates that although the detector leverages spatio-temporal information from support frames \textit{t-s} and \textit{t+s}, it fails to detect \textit{Turtle} at the target frame \textit{t}. The main reason for such missed prediction is that generated proposals from single-frame RPN overlooks \textit{Turtle} in the target frame \textit{t}. This corrupts object features during instance-level feature aggregation.
The third restriction is that these methods require several support frames to calibrate proposal feature representation for the target frame, decreasing the run time performance. Fourth, since the RPN in these methods is optimized on a single IoU level (generally IoU=0.5), they struggle to provide high-quality detections (IoU > 0.5) despite producing impressive performance on a lower IoU threshold. Moreover, these methods necessitate complex post-processing methods~\cite{chen2018optimizing, kang2017t} or additional proposal classifier networks~\cite{han2021class, han2020exploiting} to accomplish state-of-the-art performance.

To tackle the aforementioned challenges, we propose SparseVOD, an end-to-end trainable framework that exploits temporal information to learn sparse (merely 100) object proposals for VOD. 
The SparseVOD employs the recently introduced Sparse R-CNN~\cite{sun2021sparse} that has shown impressive performance by eliminating the need for dense priors enumerating over frames and alleviating the interaction between object queries and dense frame features. In particular, motivated from~\cite{wu2019sequence, gong2021temporal}, we incorporate a Temporal (Region of Interest) RoI Feature Extraction (TFE) head that replaces a single image RoI extractor in the dynamic instance interactive head in~\cite{sun2021sparse}.~Furthermore, inspired by~\cite{wu2019sequence}, we fuse attention guided Semantic Proposal Feature Aggregation (SPFA) module that enhances the feature representation of object proposals in target frames through semantic level sequence aggregation from support frames.

Contrary to most prior VOD works, our SparseVOD operates on a sparse-in sparse-out matching scheme.~This not only eliminates components like post-processing and NMS but also enables faster training network convergence without pre-training the detector as done in~\cite{he2021end}.~Furthermore, thanks to the spatio-temporal learning, iterative refinement of object proposals lead to successful predictions even on a low-quality target frame \textit{t} as shown in Fig.~\ref{fig:problem_definition}\textcolor{red}{(b)}.~Moreover, our spatio-temporal proposal learning alleviates the need for several support frames in a video and brings significant performance gains on increasing IoU thresholds (see Fig.~\ref{fig:high_quality_AP}).

Herein, our main contributions are as follows.
(1) We propose SparseVOD, a novel end-to-end trainable video object detection method. To our knowledge, \textit{this is the first work that exploits temporal information to learn object proposals for video object detection}. (2) We extend the design of Sparse R-CNN~\cite{sun2021sparse} by introducing a Temporal Feature Extraction (TFE) module that leverages temporal information to extract RoI proposal features.~Furthermore, we fuse Semantic Proposal Feature Aggregation (SPFA) in~\cite{sun2021sparse} to enhance object feature representation before final detection inspired by~\cite{wu2019sequence, gong2021temporal}.
(3) By introducing the proposed TFE and SPFA modules in~\cite{sun2021sparse}, our SparseVOD improves the baseline by far (5-6\% mAP). Without bells and whistles, our SparseVOD achieves the \textit{new best mAP of 80.3\% on the ImageNet VID benchmark using ResNet-50 as the backbone}.~Moreover, it surpasses prior state-of-the-art methods by far in terms of high-quality detections (higher IoU thresholds, see Fig.~\ref{fig:high_quality_AP}) and achieves optimal speed-accuracy tradeoff (Fig.~\ref{fig:tradeoff}).

\section{Related Work}
\label{sec:related_work}

\noindent \textbf{Proposal Learning for Image Object Detection.}
\hspace{0.8pt}
\label{subsec:image_object_detection} 
Ren \etal~\cite{ren2015faster} introduce the Region Proposal Network (RPN) in Faster R-CNN to predict object proposals. The RPN consists of a small fully convolutional network~\cite{long2015fully} that receives an anchor as an input, classifies it as an object or background, and performs box regression. This design is widely incorporated in later two-stage approaches~\cite{dai2016r, he2017mask, lin2017feature, cai2018cascade}.~MetaAnchor~\cite{yang2018metaanchor} proposes to exploit meta-learning to generate anchors dynamically. Cascade RPN~\cite{vu2019cascade} improves the object proposal quality of the conventional RPN through multi-stage refinement and adaptive convolution. Recently, Sparse R-CNN~\cite{sun2021sparse} introduces a sparse-in sparse-out paradigm that simplifies the sophisticated two-stage object detection pipeline by alleviating complex components such as Non-Maximum Suppression (NMS)~\cite{neubeck2006efficient} and dense priors.~Following a similar line of work, this paper proposes spatio-temporal learnable proposals to simplify the video object detection pipeline.



\vspace{3pt}
\noindent \textbf{Exploiting Temporal Information in Video Object Detection.}
\hspace{0.8pt}
\label{subsec:vide_object_detection}
Exploiting temporal information from other frames in a video is a natural choice to tackle the challenges of video object detection, and our work derives from the same idea. Existing approaches leveraging temporal information mainly follow one of the two directions.~The first line of works~\cite{han2016seq, kang2016object, kang2017object, sabater2020robust} mainly employ temporal information to make still-image detection results more coherent and stable. The performance of such VOD methods is sub-optimal because they are not end-to-end trainable and heavily rely on the capabilities of the initial still-image detector. On the contrary, the other direction of methods~\cite{zhu2017deep, zhu2017flow, feichtenhofer2017detect, wang2018fully, jiang2019video, wu2019sequence, han2020mining, gong2021temporal, han2021class, cui2021tf, hua2021temporal, han2020exploiting} utilizes temporal information during the course of training.~Earlier works in this category adopt FlowNet~\cite{dosovitskiy2015flownet} to propagate warp features across frames~\cite{zhu2017flow, wang2018fully, zhu2017deep, kang2017t}. However, temporal exploitation of optical flow-based works is limited to neighbouring frames, yielding inferior performance in occlusions. PSLA~\cite{guo2019progressive} proposes to learn the spatial correspondence between neighbouring frames by employing the progressive sparser stride. All these methods can only capitalize temporal information from a small number of nearby frames to refine the target frame features.~Alternatively, global feature aggregation methods~\cite{wu2019sequence, deng2019object, shvets2019leveraging} have been proposed to utilize long-term semantic information.~Recent VOD methods adopt this aggregation scheme and propose blending of temporal features~\cite{cui2021tf}, class-aware feature aggregation~\cite{han2021class, han2020exploiting}, temporal RoIAlign~\cite{gong2021temporal}, and temporal meta-adaptor~\cite{hua2021temporal} to achieve state-of-the-art results. Although these methods produce superior performance from prior efforts, their feature aggregation rely on object proposals generated without temporal information.

\vspace{3pt}
\noindent \textbf{Refining Object Proposals in Video Object Detection.}
\label{subsec:object_proposals}
\hspace{0.8pt}
Recent efforts have shown that enhancing object proposal features can alleviate the obstacles of object confusion in videos~\cite{shvets2019leveraging, han2020mining, jiang2020learning}. Shvets~\etal~\cite{shvets2019leveraging} refine the proposal for the target frame by learning similarities between proposals from different frames. LSTS~\cite{jiang2020learning} models the spatio-temporal correspondence to alleviate misalignment before aggregating features from different frames.~Han~\etal~\cite{han2020mining} propose integrating inter-video and intra-video proposals to boost target proposal features. \textit{Despite the promising improvements, the effectiveness of all these methods heavily relies on the initial quality of object proposals retrieved from single-frame RPN}~\cite{ren2015faster}.~Alternatively, this paper exploits temporal information to generate object proposals for video object detection. 

Recently, MAMBA~\cite{sun2021mamba} proposes to extract region proposals from the enhanced feature maps through a pixel-level memory bank.~TransVOD~\cite{he2021end} proposes the transformer-based VOD pipeline by extending Deformable DETR~\cite{zhu2020deformable} to exploit temporal information in videos.~Despite the simple and end-to-end trainable framework, the temporal transformer in TransVOD depends on object queries generated by the spatial transformer optimized without temporal information.~Furthermore, owing to the interaction between each object query and the dense features of an entire frame,~\cite{he2021end} is not a pure sparse method~\cite{sun2021sparse}. As a result of this dense interaction, TransVOD requires pre-training the detector on a similar dataset~\cite{lin2014microsoft}.~On the contrary, our method leverages temporal information to generate object proposals.~Moreover, following~\cite{sun2021sparse}, our proposed SparseVOD operates on a pure sparse paradigm and does not require pre-training the detector.

\vspace{-10pt}
\section{Method}
\label{sec:method}
\textbf{Overview.}~This section explains our proposed SparseVOD, which consists of two main components: Temporal RoI extraction and Semantic Proposal Feature Aggregation incorporated in the Temporal Dynamic Head. Finally, we discuss network optimization.~Note that due to space constraints, the detailed explanation of the still-image object detector Sparse R-CNN~\cite{sun2021sparse} is omitted here and can be found in the supplementary material.

\begin{figure} 
\vspace{-3pt}
\includegraphics[width=12.5cm]{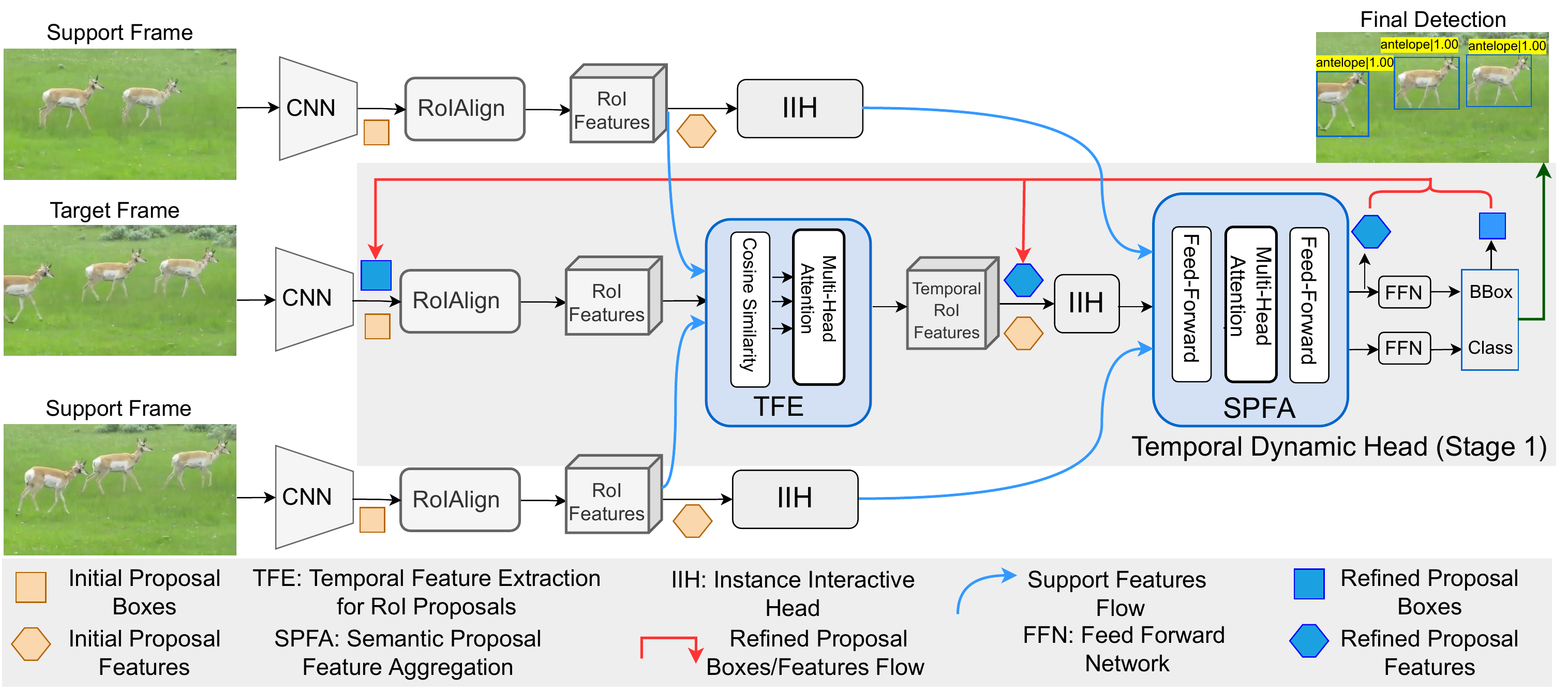}
\caption{Our SparseVOD Framework.~The input for the first stage of the temporal dynamic head consists of initial proposal boxes, proposal features, and spatial features for the target and support frames.~The TFE extracts RoI proposal features for the target frame by exploiting RoI features from support frames.~Then, SPFA receives object features from each IIH (identical as in~\cite{sun2021sparse}) and aggregates object features with guided attention heads.~This enhanced object feature representation from SPFA and box predictions after FFN serve as input proposal features and proposal boxes for the next iterative stage.}
\label{fig:pipeline}
\vspace{-16pt}
\end{figure}

\subsection{SparseVOD Architecture}
\label{subsec:sparseVOD}
The SparseVOD is a simple, end-to-end trainable framework, as shown in Fig.~\ref{fig:pipeline}. It receives a target frame and multiple support frames from the same video as input and outputs the class and location of objects in the target frame.~For each target and support frame, the extracted feature maps from the backbone, proposal boxes and corresponding proposal features are fed into the iterative Temporal Dynamic head consisting of multiple stages. We create Temporal Dynamic Head by incorporating two main components into the dynamic head used in~\cite{sun2021sparse} to effectively exploit the temporal information in videos.~First, inspired by~\cite{gong2021temporal}, we leverage support frame RoI features to extract temporal RoI features for the target frame.~Then, the corresponding Instance Interactive Head~(IIH) produces object feature representation for each video frame.~Subsequently, the Semantic Proposal Feature Aggregation module enhances the representation of the target frame by intelligently aggregating object features from support frames. The optimal object features are fed to the corresponding feed-forward network for classification and regression. Similar to~\cite{sun2021sparse}, we follow the iterative architecture in our SparseVOD. In the next iteration, the newly refined object features and the bounding box predictions serve as the target frame's proposal features and proposal boxes.

\noindent \textbf{Temporal RoI Feature Extraction.}
\label{subsec:troi}
The Sparse R-CNN~\cite{sun2021sparse} applies the conventional RoIAlign~\cite{he2017mask} pooling operation to extract proposal features, and it is widely adopted in existing VOD methods~\cite{wu2019sequence, zhu2017flow, han2020mining}. However, the naive RoIAlign operation only restricts the proposal feature extraction to exploit intra-frame features.~Therefore, motivated by~\cite{gong2021temporal}, we incorporate the Temporal Feature Extraction (TFE) module in our Temporal Dynamic Head, as illustrated in Fig.~\ref{fig:pipeline}.~The TFE leverages temporal information for the same object instance across support frames in a video. Since the features of the same object instance have high semantic resemblance across video frames, we calculate cosine similarities between target proposal features and support frame feature maps. Given target proposal features $P_{t}$ and feature map from support frame $F_{t+s}$, the cosine similarity $Sim_{t+s}(m)$ is computed as $P_{t}(m) \otimes F_{t+s}$, where $m$ represents the spatial location of $P_{t}$ and $\otimes$ denotes dot product. Note that the target frame proposal is mapped on the most similar feature maps from support frames to extract the most similar RoI features. Following the temporal attentional feature aggregation in~\cite{gong2021temporal}, we adopt the self-attention mechanism~\cite{vaswani2017attention} to aggregate the RoI features of target and support frames.

\vspace{3pt}
\noindent \textbf{Semantic Proposal Feature Aggregation.}
\label{subsec:tfa}
The Semantic Proposal Feature Aggregation (SPFA) head aims to learn the enhanced feature representation from target RoI features (containing temporal information of the same instance) and support frame RoI features to perform final classification and regression. Since we already have temporal RoI features, we follow the spirits of~\cite{wu2019sequence} and adopt semantic similarity as the metric to aggregate features from support frame proposals. We compute semantic similarity between target and support proposal features in the same way as in TFE.
For effective feature aggregation, the SPFA applies multi-head attention~\cite{vaswani2017attention} on temporal proposal features of a target frame $\overline{P_{t}}$ and support proposal features $P_{s}$ as  follows: 
\vspace{-6pt}
\begin{equation}
\vspace{-6pt}
    W_t = softmax( \frac{ \phi(P_{s}) \cdot (\theta(\overline{P_{t}}))^T }{\sqrt{d_{\theta(\overline{P_{t}})}}})\cdot \sigma(P_{s})
\end{equation}
\noindent where $\phi(.)$, $\theta(.)$, and $\sigma(.)$ are some linear transformations. The symbol $T$ represents transposition, $d$ denotes the size of transformed $\overline{P_{t}}$, and $W_t$ is the enhanced proposal representation of a target frame. This rich instance representation improves robustness against inherent challenges of VOD, such as appearance deterioration.

\vspace{3pt}
\noindent \textbf{Loss Function.}
\label{subsec:loss_function}
Since our SparseVOD operates on a one-to-one label matching, our method's loss function and training process are similar to the original Sparse R-CNN. We adopt set predictions loss~\cite{carion2020end, zhu2020deformable, he2021end}, which aims to optimize the bipartite matching among the ground truth and predictions. Following~\cite{he2021end, sun2021sparse, carion2020end}, the cost function is defined as $\mathcal{L} = \lambda_{cls} \cdot \mathcal{L}_{cls}+ \lambda_{L1} \cdot \mathcal{L}_{L1} + \lambda_{giou} \cdot \mathcal{L}_{giou}$, where $\mathcal{L}_{cls}$ is the focal loss~\cite{lin2017focal} for classification. $\mathcal{L}_{L1}$ and $\mathcal{L}_{giou}$ are L1 loss and  generalized IoU loss~\cite{rezatofighi2019generalized} for regression, respectively. $\lambda_{cls}$, $\lambda_{L1}$, and $\lambda_{giou}$ are coefficients to balance the loss. We employ an identical setting to balance these losses as in~\cite{sun2021sparse}.
\vspace{-15pt}

\section{Experiments and Results}
\label{sec:experiments}
\vspace{-5pt}
\subsection{Experimental Settings}
\label{subsec:experimental setup}
We perform experiments on the ImageNet VID dataset~\cite{russakovsky2015imagenet}, which comprises 3862 training videos and 555 validation videos. Following prior works~\cite{cui2021tf, wu2019sequence, zhu2017flow}, we train our model on a combination of ImageNet VID and DET datasets and evaluate the results on the validation set. We adopt ImageNet pre-trained~\cite{deng2009imagenet} ResNet-50~\cite{he2016deep}, ResNet-101, and ResNeXt-101~\cite{xie2017aggregated} backbones to compare performance with recent state-of-the-art methods. We train our network for 12 epochs with a batch size of 8 on 8 GPUs. Analogous to~\cite{sun2021sparse}, we use AdamW~\cite{loshchilov2017decoupled} optimizer with a weight decay of $10^{-4}$.  Initially, the learning rate is set to $2.5\times10^{-5}$ and divided by 10 at the 8-th and 11-th epochs. Following~\cite{sun2021sparse, carion2020end, zhu2020deformable}, we set $\lambda$\textsubscript{cls}=2, $\lambda$\textsubscript{L1}=5, and $\lambda$\textsubscript{giou}=2. We follow the basic settings of~\cite{sun2021sparse} and set the number of iterative stages, proposal boxes, and the corresponding proposal features to 6, 100, and 100, respectively. We refer readers to supplementary materials for more details. 

\subsection{Main Results}
\label{subsec:qantity}
We compare the performance of the proposed SparseVOD with prior state-of-the-art VOD methods on ImageNet VID dataset in Table~\ref{table:sota_comparison}.~Besides the conventional mAPs @IoU=0.5, we compute mAPs @IoU=0.75 and @IOU=0.5:95 as in~\cite{lin2014microsoft} to analyze the precision of detections. Owing to the unavailability of code at the time of experiments, apart from~\cite{he2021end, sun2021mamba}, we reproduce the results of existing methods from the code provided by the original papers for direct comparison. It is important to mention that all the results shown in Table~\ref{table:sota_comparison} are without any post-processing. By looking at results under the backbone of ResNet-50, it is evident that our SparseVOD outperforms recent methods~\cite{zhu2017flow, wu2019sequence, chen2020memory, gong2021temporal}, mainly relying on feature aggregation of region proposals.~Furthermore, it surpasses the previous best score of 79.9\% by~\cite{he2021end} and achieves a new best score of 80.3\% on ResNet-50.~Note that alongside the improvement on mAP @IoU=0.5, our SparseVOD demonstrates a significant increase (7.3 and 6 points) in the precise localization on mAPs @IoU=0.75 and @IOU=0.5:95 from the previous best method~\cite{gong2021temporal}, reflecting the superiority of spatio-temporal learnable proposals. 

\begin{table}
\begin{center}
\scriptsize
\begin{tabular}{ccccccc}
\toprule
Methods & Venue & Backbone & Detector & mAP$_{50:95}$($\%$) & mAP$_{50}$($\%$)  &  mAP$_{75}$($\%$)\\
\noalign{\smallskip}
\hline
FGFA*~\cite{zhu2017flow}  & ICCV'17 & ResNet-50 & Faster R-CNN & 47.1 & 74.7 & 52.0 \\
SELSA*~\cite{wu2019sequence}  & ICCV'19 & ResNet-50 & Faster R-CNN & 48.6 & 78.4 & 52.5 \\
MEGA*~\cite{chen2020memory}  & CVPR'20 & ResNet-50 & Faster R-CNN & 48.1 & 77.3 & 52.2 \\
TROI*~\cite{gong2021temporal}  & AAAI'21 & ResNet-50 & Faster R-CNN & \textbf{\color{blue} 48.8}  & 78.9 & \textbf{\color{blue}52.8} \\
TransVOD~\cite{he2021end}  & ACM MM'21 & ResNet-50 & Deformable DETR& -  & \textbf{\color{blue}79.9} & - \\
\hline
{Frame Baseline~\cite{sun2021sparse}}& CVPR'21 & ResNet-50 & Sparse R-CNN &48.7 & 71.1 & 52.4 \\
{\textbf{SparseVOD} }& BMVC'22 & ResNet-50 & Sparse R-CNN & \textbf{\color{red}54.7} & \textbf{\color{red}80.3} & \textbf{\color{red}60.1} \\
\hline
FGFA*~\cite{zhu2017flow}  & ICCV'17 & ResNet-101 & Faster R-CNN & 50.4 & 78.1 & 56.7 \\
SELSA*~\cite{wu2019sequence}  & ICCV'19 & ResNet-101 & Faster R-CNN & 52.4 & 81.5 &  57.9 \\
MEGA*~\cite{chen2020memory}  & CVPR'20 & ResNet-101 & Faster R-CNN & \textbf{\color{blue}53.1}  & \textbf{\color{blue} 82.9} & \textbf{\color{blue}59.1}\\
TROI*~\cite{gong2021temporal}  & AAAI'21 & ResNet-101 & Faster R-CNN& 51.6  &   82.6 & 56.3 \\
MAMBA~\cite{sun2021mamba}  & AAAI'21 & ResNet-101 & Faster R-CNN& -  &  \textbf{\color{red} 84.6} & - \\
TransVOD~\cite{he2021end}  & ACM MM'21 & ResNet-101 & Deformable DETR & - &81.9 & - \\
\hline
{Frame Baseline~\cite{sun2021sparse}}& CVPR'21 & ResNet-101 & Sparse R-CNN & 51.7 & 74.6 & 53.9 \\
{\textbf{SparseVOD} }& BMVC'22 & ResNet-101 & Sparse R-CNN  & \textbf{\color{red}56.9} & 81.9 & \textbf{\color{red}63.1}\\
\hline
FGFA*~\cite{zhu2017flow}  & ICCV'17 & ResNeXt-101 & Faster R-CNN& 52.5  & 79.6 & 59.8 \\
SELSA*~\cite{wu2019sequence}  & ICCV'19 & ResNeXt-101 & Faster R-CNN& \textbf{\color{blue}  54.2} & 83.1 &  \textbf{\color{blue} 61.3}  \\
MEGA~\cite{chen2020memory}  & CVPR'20 & ResNeXt-101 & Faster R-CNN  & - & \textbf{\color{blue}84.1}& -\\
TROI*~\cite{gong2021temporal}  & AAAI'21 & ResNeXt-101 & Faster R-CNN  & 54.4 & \textbf{\color{red} 84.3} & 60.9\\
\hline
{Frame Baseline~\cite{sun2021sparse}}& CVPR'21 & ResNeXt-101 & Sparse R-CNN & 53.3 & 76.6 & 57.9 \\
{\textbf{SparseVOD} }& BMVC'22 & ResNeXt-101 & Sparse R-CNN  & \textbf{\color{red}58.0} & 83.1 & \textbf{\color{red}64.3}\\
\bottomrule
\end{tabular}
\end{center}
\caption{Comparison with other state-of-the-art methods on the ImageNet VID dataset. Results with * are reproduced. The two best results are highlighted in red and blue.}
\label{table:sota_comparison}
\vspace{-20pt}
\end{table}

When stronger backbones of ResNet-101 and ResNeXt-101 are incorporated into SparseVOD, the performance (mAP$_{50}$) further increases to 81.9\% and 83.1\%, respectively. Note that although MAMBA~\cite{sun2021mamba}, MEGA~\cite{chen2020memory}, and TROI~\cite{gong2021temporal} demonstrate better results at mAP$_{50}$, our SparseVOD supersedes them by far (4$\sim$5 points in mAP) on higher IoU thresholds. These results correspond to our argument that while prior VOD methods operating on dense to sparse detection pipelines show impressive results on mAP$_{50}$, they fail to produce confident and precise predictions. Furthermore, our SparseVOD boosts the single-frame baseline (Sparse R-CNN~\cite{sun2021sparse}) by a strong margin (8\%$\sim$9\% mAP$_{50}$) with all backbone networks. This noticeable increase in mAP highlights the importance of leveraging temporal information to generate object proposals in VOD.
\vspace{-5pt}
\subsection{Iterative proposal Visualization Analysis}
\label{subsec:qualitative}
\begin{figure}[H]
\centering
\vspace{-10pt}
\includegraphics[width=11.5cm]{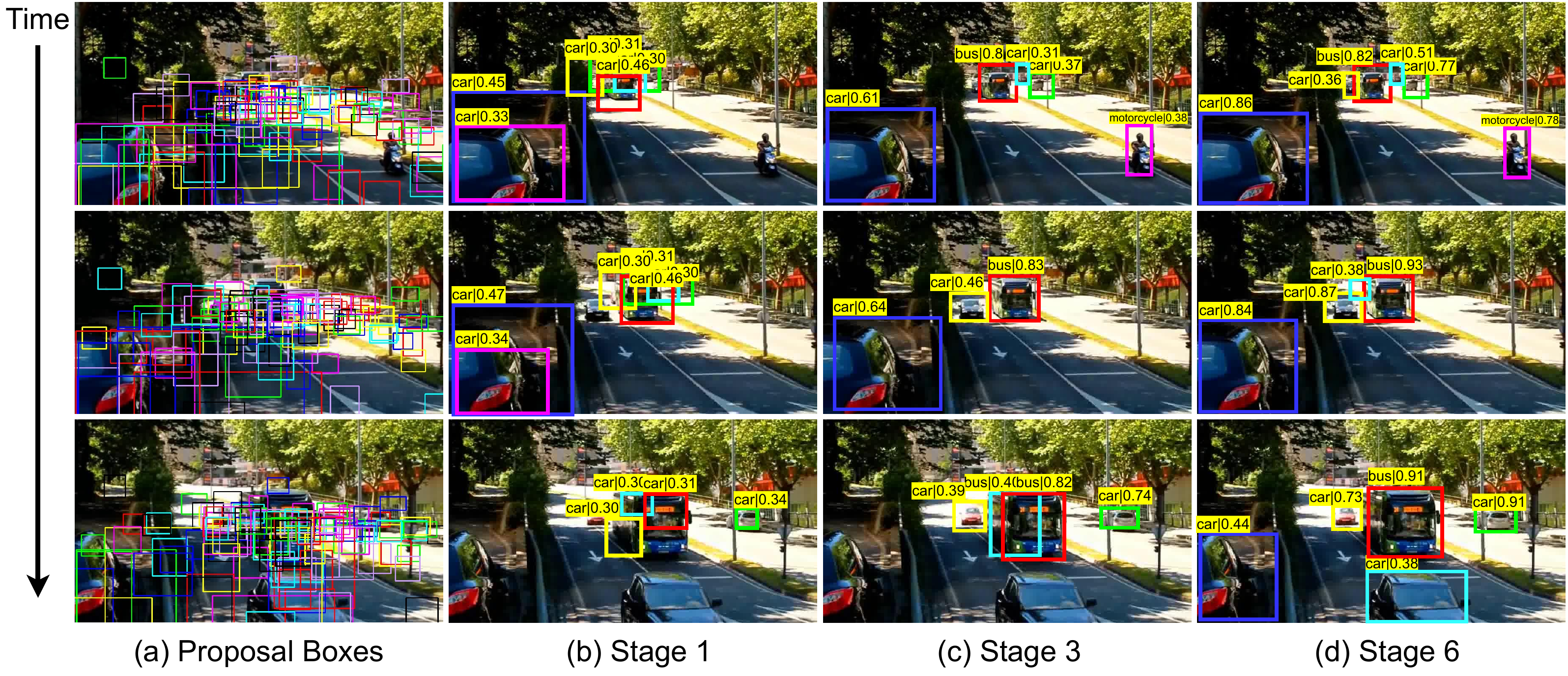}
\vspace{-5pt}
\caption{Illustration of the learned proposals and bounding box predictions at different iterative stages from a converged model. For brevity, we only visualize predictions with a confidence score greater than 0.3.~Note that learned proposal boxes in (a) cover possible regions in all three video frames while the cascading heads in each stage enhance detections.}
\label{fig:iterative_results}
\vspace{-20pt}
\end{figure}

\vspace{-5pt}
We visualize the behaviour of learned proposals boxes of a trained model on a video clip from the validation set in Fig.~\ref{fig:iterative_results}.~Note that these proposals cover almost all potential regions in the target frames.~This ensures high recall performance even with sparse proposals.~Moreover, each stage in the cascaded architecture refines the bounding box offset and removes duplication.~This makes our pipeline independent of any post-processing techniques to produce precise predictions.~Fig.~\ref{fig:iterative_results}\textcolor{red}{(d)} further exhibits the robustness of our SparseVOD by producing high-quality predictions in challenging scenarios with camera defocus and part-occlusions. Please see supplementary materials for more qualitative analysis. \vspace{-5pt}

\subsection{Ablation Studies}
\label{subsec:ablation}
\vspace{-5pt}
This section discusses the effect of key components and validates the design choices in our proposed method. Following~\cite{he2021end}, we perform all experiments on ImageNet VID dataset with ResNet-50 as the backbone. The run time (FPS) is tested on a single DGX A100 GPU. More ablation studies can be found in supplementary materials.

\vspace{3pt}
\noindent \textbf{Effectiveness of each component in SparseVOD.} 
\hspace{1pt}
Table~\ref{table:ablation_each_module} summarizes the impact of adding each component to build our proposed method. Beginning with the single-frame baseline, we incorporate the TFE module (explained in Section~\ref{subsec:troi}) and boost the AP$_{50}$ from 71.1\% to 76.9\%.~This demonstrates the benefit of exploiting inter-frame information to extract RoI proposal features. On the other hand, by separately plugging the SPFA module (explained in Section~\ref{subsec:tfa}) into a single-frame baseline, we achieve substantial gains in AP$_{50}$ from 71.1\% to 79.1\%. Finally, by combining both TFE and SPFA to build the proposed SparseVOD, we gain a further boost of 1.2\% in AP$_{50}$, accomplishing 80.3\%. These results establish the superiority of introducing spatio-temporal feature aggregation for learnable proposals.
\newcommand{\cmark}{\ding{51}}%
\newcommand{\xmark}{\ding{55}}%
\begin{table}[H]
\small
\begin{center}
\begin{tabular}{ccc|cccc}
\toprule
Single Frame Baseline & TFE & SPFA  & AP$_{50}$(\%) &  AP$_{75}$($\%$)  &  AP$_{50:95}$($\%$)& FPS \\
\hline\noalign{\smallskip}
\textcolor{Green}{\cmark} & \color{red}\xmark & \color{red}\xmark  & 71.1 & 52.4 & 48.7 &  24.3 \\
 \color{Green}\cmark &  \color{Green}\cmark&  \color{red}\xmark & 76.9$_{\uparrow}$\textsubscript{5.8} & 57.5$_{\uparrow}$\textsubscript{5.1} &52.1$_{\uparrow}$\textsubscript{3.4} & 14.9 \\
 \color{Green}\cmark & \color{red}\xmark & \color{Green}\cmark& 79.1$_{\uparrow}$\textsubscript{8.0} & 59.0$_{\uparrow}$\textsubscript{6.6} & 54.1$_{\uparrow}$\textsubscript{5.4}& 17.7  \\
 \color{Green}\cmark &  \color{Green}\cmark& \color{Green}\cmark& 80.3$_{\uparrow}$\textsubscript{9.2} & 60.1$_{\uparrow}$\textsubscript{7.7} & 54.7$_{\uparrow}$\textsubscript{6.0} &  14.4 \\
\bottomrule
\end{tabular}
\end{center}
\vspace{-5pt}
\caption{Ablation on effectiveness of each module in SparseVOD.}
\label{table:ablation_each_module}
\vspace{-15pt}
\end{table}
\begin{minipage}{\linewidth}
  \begin{minipage}{0.35\linewidth}
    \vspace{-10pt}
    \hspace{-10pt}
  \begin{table}[H]
    \small
    \begin{tabular}{c|ccc}
        \toprule
        Stages & AP$_{50}$(\%) & AP$_{75}$(\%) & FPS\\
        \hline
        1 & 60.0 & 27.2 & 42.2\\
        2 & 74.5 & 52.8 & 29.4\\
        3 & 78.0 & 58.3 & 23.7\\
        4 & 78.5 & 58.6 & 19.8\\
        5 & 79.1 & 59.8 & 16.1\\
        {6} & \textbf{80.3} & \textbf{60.1} & \textbf{14.4}\\
        12 & 77.1 & 55.2 & 5.7\\
        \bottomrule
  \end{tabular}\\
  \vspace{-5pt}
  \caption{Ablation on the number of stages.}
  \label{subtab:stages}
  \end{table}
  \end{minipage}
  \hspace{0.05\linewidth}
  \begin{minipage}{0.5\linewidth}
  \vspace{-10pt}

      \begin{figure}[H]
        \includegraphics[width=7cm]{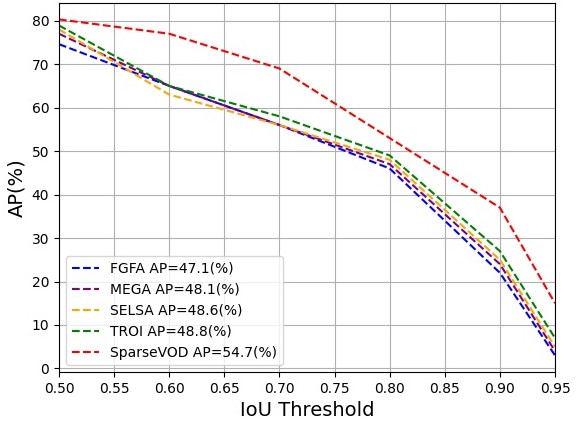}
        \vspace{-20pt}
        \caption{The detection performance of recent VOD methods and our SparseVOD on increasing IoU threshold.}
        \label{fig:high_quality_AP}
    \end{figure}
  \end{minipage}
\end{minipage}

\begin{table}[H]
\footnotesize
\begin{center}
\begin{tabular}{cccc|cccc}
\toprule\noalign{\smallskip}
TFE (Avg.) & TFE (MSA) & SPFA (Avg.) & SPFA (MSA)  & AP$_{50}$(\%) &  AP$_{75}$($\%$)  &  AP$_{50:95}$($\%$)& FPS \\
\hline \noalign{\smallskip}
     \color{red}\xmark & \color{red}\xmark & \color{red}\xmark 
     & \color{red}\xmark  &71.1 & 52.4 & 48.7 &  24.3 \\
     \hline
     \color{Green}\cmark & \color{red}\xmark & \color{Green}\cmark 
     & \color{red}\xmark  &75.8 & 52.9 & 49.1 &  18.5 \\
     
     \color{Green}\cmark & \color{red}\xmark & \color{red}\xmark  & \color{Green}\cmark &79.4 & 58.9 & 54.1 &  16.3 \\
 
    \color{red}\xmark  &\color{Green}\cmark &  \color{Green}\cmark & \color{red}\xmark  &78.5 & 58.1 & 52.1 &  14.9 \\
    
    \color{red}\xmark  &\color{Green}\cmark & \color{red}\xmark  & \color{Green}\cmark & \textbf{80.3} & \textbf{60.1} & \textbf{54.7} & 14.4\\
\bottomrule
\end{tabular}
\end{center}
\caption{Ablation on the effectiveness of multi-head attention in TFE (Temporal Feature Extraction) and SPFA (Semantic Proposal Feature Aggregation) modules.~The first row represents results from a single-frame baseline. The terms Avg.~and MSA denote simple averaging and multi-head self-attention.}
\label{table:ablation_attention}
\vspace{-20pt}
\end{table}
    
\begin{minipage}{\linewidth}

  \begin{minipage}{0.35\linewidth}
  \begin{table}[H]
  \hspace{-10pt}
    \small
        \begin{tabular}{c|ccc}
            \toprule
            N\textsubscript{ref} & AP$_{50}$(\%) &  AP$_{75}$($\%$)  & FPS \\
            \hline
            1 & 71.1 & 52.4 & 24.3\\
            2 & 79.0 & 58.2 & 17.7\\
            4 & 79.9 & 59.5 & 15.9\\
            {6} & \textbf{80.3} & \textbf{60.1} & \textbf{14.4}\\
            10 & 80.2 & 60.2 & 11.7\\
            14 & 80.3 & 60.2 & 9.5\\
            \bottomrule
        \end{tabular}\\ 
        \vspace{-5pt}
          \caption{Ablation on number of support frames.}
          \label{tab:ablation_support_frames}
      \end{table}
  \end{minipage}
  \hspace{0.03\linewidth}
  \begin{minipage}{0.55\linewidth}
        \begin{figure}[H]
        \includegraphics[width=7cm]{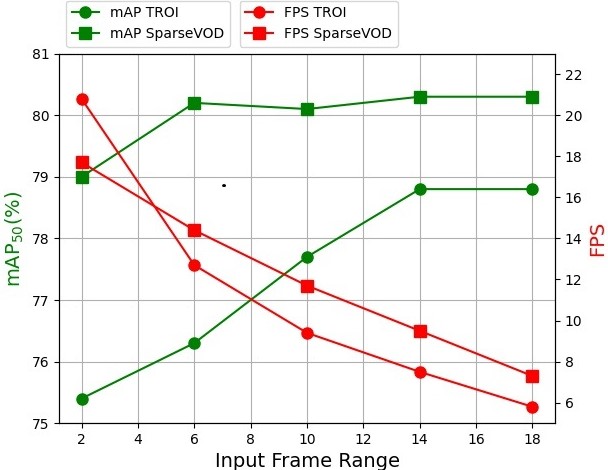}
        \vspace{-10pt}
        \caption{Speed-accuracy tradeoff between SparseVOD and previous best competitor (TROI+SELSA~\cite{gong2021temporal}).}
        \label{fig:tradeoff}
      \end{figure}
  \end{minipage}
\end{minipage}

\vspace{5pt}
\noindent \textbf{Number of Stages.}
The impact of increasing stages in an iterative architecture is summarized in Table~\ref{subtab:stages}. Note that without iterative architecture, even though the performance AP$_{50}$(\%) reaches 60.0, the AP$_{75}$ is merely 27.2\%. Since the input proposal boxes at the first stage are just random distribution of possible object locations, this result (AP$_{50}$=60.0\%) indicates that computing AP on a single IoU threshold of 0.5 is not a reliable evaluation metric.~By increasing iterative stages to 3, the performance already reaches a comparable AP$_{50}$ of 78\% and surpasses prior methods on AP$_{75}$ with 58.3\%. Finally, similar to~\cite{sun2021sparse}, the performance saturates at 6 stages. Hence, we adopt 6 stages in our experiments.

\vspace{2pt}
\noindent \textbf{Comparing High-Quality Detection.}
Fig.~\ref{fig:high_quality_AP} shows the AP curves of recent state-of-the-art VOD methods and our SparseVOD under increasing IoU thresholds. It is evident that the proposed method consistently outperforms prior works with a significant margin on all the evaluation metrics.~Note that although the difference is mild with the previous best competitor TROI~\cite{gong2021temporal}, on low IoU threshold (0.5), it rises on higher IoU thresholds. These results reflect the superiority of sparse spatio-temporal learnable proposals over hand-crafted dense priors optimized on a single IoU level in existing VOD methods.
\vspace{2pt}

\noindent \textbf{Effectiveness of Multi-head Attention in TFE and SPFA.}
We demonstrate the effectiveness of multi-head attentional blocks~\cite{vaswani2017attention} in Temporal Feature Extraction~(TFE) and Semantic Proposal Feature Aggregation~(SPFA) modules in Table~\ref{table:ablation_attention}.~For direct comparison, we conduct baseline experiments where attentional weights in TFE and SPFA are replaced by simple averaging to aggregate features from target and support frames. As shown in Table~\ref{table:ablation_attention}, with averaging in both modules, the AP\textsubscript{50} reaches 75.8\%, reflecting the benefit of leveraging temporal information to refine object proposals. Furthermore, we conduct experiments by switching attention to averaging in one of the two modules. With TFE~(Avg) and SPFA~(MSA), we observe an impressive speed-accuracy tradeoff of 79.4\% AP\textsubscript{50} and 16.3 FPS. Since the averaging in TFE is performed on the most similar RoI features computed with cosine similarities, when combined with SPFA~(MSA), it already provides an acceptable object feature representation. However, these results are still inferior (-0.9 points in AP\textsubscript{50}) to the performance achieved when multi-head attention is plugged in both TFE and SPFA. These results (80.3\% AP\textsubscript{50}) indicate the superiority of multi-head attentional aggregation in our method.

\noindent \textbf{Number of Support Frames.}
Table~\ref{tab:ablation_support_frames} presents the ablations on the number of support frames. We follow the identical frame sampling strategy as in~\cite{wu2019sequence, gong2021temporal}, where support frames are uniformly sampled from the entire video. We can see that the AP\textsubscript{50} already reaches 79.0\% with 2 support frames. Upon increasing support frames, the performance keeps increasing and tends to stabilize after reaching the AP\textsubscript{50} of 80.3\% with 6 support frames.

\noindent \textbf{Speed-accuracy Tradeoff.}
\label{subsec:comp_analysis}
Table~\ref{table:ablation_each_module} shows that the computational load in our SparseVOD stems from Temporal Feature Extraction~(TFE) and Semantic Proposal Feature Aggregation~(SPFA). Since the results of~\cite{he2021end} are not reproducible, for direct comparison, we analyse the speed-accuracy tradeoff of the second best method TROI~\cite{gong2021temporal} and our SparseVOD in Fig.~\ref{fig:tradeoff}.~Note that TROI~\cite{gong2021temporal} is built upon SELSA~\cite{wu2019sequence} to enhance performance.~With only 6 support frames sampled from the entire video, our SparseVOD achieves a new best AP\textsubscript{50} of 80.3\% with a run time of 14.4 FPS. In contrast, TROI manages to reach its best performance (AP\textsubscript{50} of 78.8\%) with a run time of 7.5 FPS after utilizing 14 support frames. The results in Fig.~\ref{fig:tradeoff} demonstrate that the temporal feature aggregation from several support frames in prior works~\cite{wu2019sequence, gong2021temporal} lead to a major increase in run time, producing a sub-optimal speed-accuracy tradeoff. Contrarily, thanks to the spatio-temporal learnable proposals, our SparseVOD yields an optimal speed-accuracy tradeoff~(79.0\% AP\textsubscript{50} and 17.7 FPS) with merely 2 support frames.


\vspace{-15pt}
\section{Conclusion}
\label{sec:conclusion}
\vspace{-5pt}
This paper proposes SparseVOD, a novel video object detection pipeline which introduces spatio-temporal feature aggregation to refine object proposals.~The SparseVOD effectively eliminates hand-crafted dense priors and provides reliable proposal features even with deteriorated input frames.~Particularly, the SparseVOD incorporates attention-guided Temporal Feature Extraction and Semantic Proposal Feature Aggregation modules in Sparse R-CNN~\cite{sun2021sparse}. Extensive experiments validate that SparseVOD significantly improves the baseline performance by 8\%-9\% in mAP and achieves the state-of-the-art 80.3\% mAP\textsubscript{50} on the ImageNet VID dataset with ResNet-50 backbone.~Besides, our SparseVOD beats existing methods in terms of high-quality predictions and optimal speed-accuracy tradeoff.~To our knowledge, our work is the first one that exploits temporal information in directly generating a sparse set of object proposals for video object detection.~We hope similar work can be applied to other video analysis tasks like object tracking and video instance segmentation.

\bibliography{references}
\end{document}